\documentclass[11pt, letterpaper, oneside]{article}
\usepackage[a4paper,left=2.5cm,right=2.5cm,top=2.5cm,bottom=2.5cm]{geometry}

\usepackage{graphicx}
\usepackage{amsmath}
\usepackage{amssymb}
\usepackage{mathtools}
\usepackage{xcolor}
\usepackage{hyperref}

\begin{document}
\title{ECQ$^{\text{x}}$: Explainability-Driven Quantization for Low-Bit and Sparse DNNs}
\author{Daniel Becking$^\text{1}$\and
Maximilian Dreyer$^\text{1}$ \and
Wojciech Samek$^{\text{1,2,}\dagger}$\and
Karsten Müller$^{\text{1,}\dagger}$\and
Sebastian Lapuschkin$^{\text{1,}\dagger}$\and \\
{\small $^{\text{1}}$ Department of Artificial Intelligence, Fraunhofer Heinrich Hertz Institute, Berlin, Germany} \\
{\small $^{\text{2}}$ BIFOLD – Berlin Institute for the Foundations of Learning and Data, Berlin, Germany} \\
{\small $^\dagger$ \texttt{\{wojciech.samek,karsten.mueller,sebastian.lapuschkin\}@hhi.fraunhofer.de}}
}
\date{}
\maketitle

\begin{abstract}
The remarkable success of deep neural networks (DNNs) in various applications is accompanied by a significant increase in network parameters and arithmetic operations.
Such increases in memory and computational demands make deep learning prohibitive for resource-constrained hardware platforms such as mobile devices.
Recent efforts aim to reduce these overheads, while preserving model performance as much as possible, and include parameter reduction techniques, parameter quantization, and lossless compression techniques. 

In this chapter, we develop and describe a novel quantization paradigm for DNNs:
Our method leverages concepts of explainable AI (XAI) and concepts of information theory:
Instead of assigning weight values based on their distances to the quantization clusters,
the assignment function additionally considers weight relevances obtained from Layer-wise Relevance Propagation (LRP) and the information content of the clusters (entropy optimization).
The ultimate goal is to preserve the most relevant weights in quantization clusters of highest information content.\\
Experimental results show that this novel Entropy-Constrained and XAI-adjusted Quantization (ECQ$^{\text{x}}$) method generates ultra low-precision (2-5 bit)
and simultaneously sparse neural networks while maintaining or even improving model performance.
Due to  reduced parameter precision and high number of zero-elements,
the rendered networks are highly compressible in terms of file size,
up to $103\times$ compared to the full-precision unquantized DNN model.
Our approach was evaluated on different types of models and datasets (including Google Speech Commands, CIFAR-10 and Pascal VOC) and compared with previous work.
\end{abstract}

\section{Introduction}

Solving increasingly complex real-world problems, continuously contributes to the success of deep neural networks (DNNs)\cite{schutt2017quantum,senior2020improved}.
DNNs have long been established in numerous machine learning tasks and for this have been significantly improved in the past decade. This is often achieved by over-parameterizing models, i.e., their performance is attributed to their growing topology, adding more layers and parameters per layer~\cite{simonyan2014very,he2016deep}. Processing a very large number of parameters comes at the expense of memory and computational efficiency. The sheer size of state-of-the-art models makes it difficult to execute them on resource-constrained hardware platforms. In addition, an increasing number of parameters implies higher energy consumption and increasing run times. 

Such immense storage and energy requirements however contradict the demand for efficient deep learning applications for an increasing number of hardware-constrained devices,
e.g., mobile phones, wearable devices, Internet of Things, autonomous vehicles or robots.
Specific restrictions of such devices include limited energy, memory, and computational budget. Beyond these, typical applications on such devices, e.g., healthcare  monitoring, speech recognition, or autonomous driving, require low latency and/or data privacy. These latter requirements are addressed by executing and running the aforementioned applications directly on the respective devices (also known as ``edge computing'') instead of transferring data to third-party cloud providers prior to processing.

In order to tailor deep learning to resource-constrained hardware, a large research community has emerged in recent years~\cite{SH_survey,warden2020tinyml}.
By now, there exists a vast amount of tools to reduce the number of operations and model size, as well as tools to reduce the precision of operands and operations (bit width reduction, going from floating point to fixed point). Topics range from neural architecture search (NAS), knowledge distillation, pruning/sparsification, quantization, lossless compression and hardware design.

Beyond all, quantization and sparsification are very promising and show great improvements in terms of neural network efficiency optimization~\cite{hoefler2021sparsity,VS_survey}. Sparsification sets less important neurons or weights to zero and quantization reduces  parameter's bit widths from default 32 bit float to, e.g., 4 bit integer. These two techniques enable higher computational throughput, memory reduction and skipping of arithmetic operations for zero-valued elements, just to name a few benefits.
However, combining both high sparsity and low precision is challenging, especially when relying only on the weight magnitudes as a criterion for the assignment of weights to quantization clusters.

In this work, we propose a novel neural network quantization scheme to render low-bit \textit{and} sparse DNNs. More precisely, our contributions can be summarized as follows:

\begin{enumerate}
    \item Extending the state-of-the-art concept of entropy-constrained quantization (ECQ) to utilize concepts of XAI in the clustering assignment function.\\ 
    \item Use relevances observed from Layer-wise Relevance Propagation (LRP) at the granularity of per-weight decisions to correct the magnitude-based weight assignment.\\
    \item Obtaining state-of-the-art or better results in terms of the trade-off between efficiency and performance compared to the previous work.
\end{enumerate}

The chapter is organized as follows: First, an overview of related work is given.
Second, in Section~\ref{sec:nn-quant}, basic concepts of neural network quantization are explained, followed by entropy-constrained quantization. Section \ref{sec:XQ} describes the ECQ extension towards ECQ$^{\text{x}}$ as an explainability-driven approach. Here, LRP is introduced and the per-weight relevance derivation for the assignment function presented. Next, the ECQ$^{\text{x}}$ algorithm is described in detail. Section \ref{sec:exp} presents the experimental setup and obtained results, followed by the final conclusion in Section \ref{sec:conc}.

\section{Related Work}

A large body of literature exists that has focused on improving DNN model efficiency.
Quantization  is  an  approach  that  has  shown  great  success \cite{quant_survey}. While most research focuses on reducing the bit width for inference, \cite{zhou2016dorefa} and others focus on quantizing weights, gradients and activations to also accelerate backward pass and training. Quantized models often require fine-tuning or re-training to adjust model parameters and compensate for quantization-induced accuracy degradation. This is especially true for precisions $<8$ bit (cf. Figure~\ref{weights_vs_acts} in Section~\ref{sec:nn-quant}). Trained quantization is often referred to as ``quantization-aware training'', for which additional trainable parameters may be introduced (e.g., scaling parameters \cite{Bhalgat_2020_CVPR_Workshops} or directly trained quantization levels (centroids) \cite{TTT}). A precision reduction to even 1 bit was introduced by BinaryConnect \cite{courbariaux2015binaryconnect}. However, this kind of  quantization usually results in severe accuracy drops. As an extension, ternary networks allow weights to be zero, i.e., constraining them to 0 in addition to $w_{-}$ and $w_{+}$, which yields results that outperform the binary counterparts \cite{EC2T}.
In DNN quantization, most clustering approaches are based on distance measurements between the unquantized weight distribution and the corresponding centroids. The works in \cite{entropyChoiEL16} and \cite{entropyPark17} were pioneering in using Hessian-weighted and entropy-constrained clustering techniques. To the best of our knowledge, \cite{deeplift} were the first and only ones to use concepts of XAI for DNN quantization. They use DeepLIFT importance measures which are restricted to the granularity of convolutional channels, whereas our proposed ECQ$^{\text{x}}$ computes LRP relevances per weight.

Another method for reducing the memory footprint and computational cost of DNNs is sparsification. In the scope of sparsification techniques, weights with small saliency (i.e., weights which minimally affect the model's loss function) are set to zero, resulting in a sparser computational graph and higher compressible matrices. Thus, it can be interpreted as a special form of quantization, having only one quantization cluster with centroid value 0 to which part of the parameter elements are assigned to.
This sparsification can be carried out as unstructured sparsification \cite{SH_pruning}, where any weight in the matrix with small saliency is set to zero, independently of its position.
Alternatively, a structured  sparsification is applied, where an entire regular subset of parameters is set to zero,
e.g., entire  convolutional  filters, matrix rows or columns \cite{he2017channel}. ``Pruning'' is conceptually related to sparsification but actually removes the respective weights rather than setting them to zero.
This  has the effect of changing the number of input and output shapes of layers and weight matrices\footnote{In practice, pruning is often simulated by masking, instead of actually restructuring the model's architecture.}. 
Most pruning/ sparsification approaches are magnitude-based, i.e., weight saliency is approximated by the weight values, which is straightforward. However, since the early 1990s methods that use, e.g., second-order Taylor information for weight saliency \cite{lecun1990optimal} have been used alongside other criteria ranging from random pruning to correlation and similarity measures (for the interested reader we recommend \cite{hoefler2021sparsity}). In \cite{yeom2021pruning}, LRP relevances were first used for structured pruning.

Generating efficient neural network representations can also be a result of combining multiple techniques. In Deep Compression~\cite{DeepCompression}, a three-stage model compression pipeline is described. First, redundant connections are pruned iteratively. Next, the remaining weights are quantized. Finally, entropy coding is applied to further compress the weight matrices in a lossless manner. This three stage model is also used in the new international ISO/IEC standard on Neural Network compression and Representation (NNR)~\cite{NNROverview}, where efficient data reduction, quantization and entropy coding methods are combined. For coding, the  highly efficient universal  entropy  coder DeepCABAC \cite{DeepCABAC} is used, which yields compression gains  of up  to $63\times$.
Although the proposed method achieves high compression gains, the compressed representation of the DNN weights require decoding prior to performing inference.
In contrast, compressed matrix formats like Compressed  Sparse  Row (CSR) derive a representation that enables inference directly in the compressed format \cite{CER}.

Orthogonal to the previously described approaches is the research area of Neural Architecture Search (NAS)\cite{NAS_survey}.
Both manual \cite{sandler2018mobilenetv2} and automated \cite{tan2019mnasnet} search strategies have played an important role in optimizing DNN architectures in terms of latency, memory footprint, energy consumption, etc.
Microstructural changes include, e.g., the replacement of standard convolutional layers by more efficient types like depth-wise or point-wise convolutions, layer decomposition or factorization, or kernel size reduction.
The macro architecture specifies the type of modules (e.g., inverted residual), their number and connections.

Knowledge distillation (KD) \cite{Hinton2015DistillingTK} is another active branch of research that aims at generating efficient DNNs. The KD paradigm leverages a large teacher model that is used to train a smaller (more efficient) student model. Instead of using the “hard” class labels to train the student, the key idea of model distillation is to deploy the teacher's class probabilities, as they can contain more information about the input.
\section{Neural Network Quantization}
\label{sec:nn-quant}
\begin{figure}[t]
\centering
\includegraphics[width=1.0\textwidth]{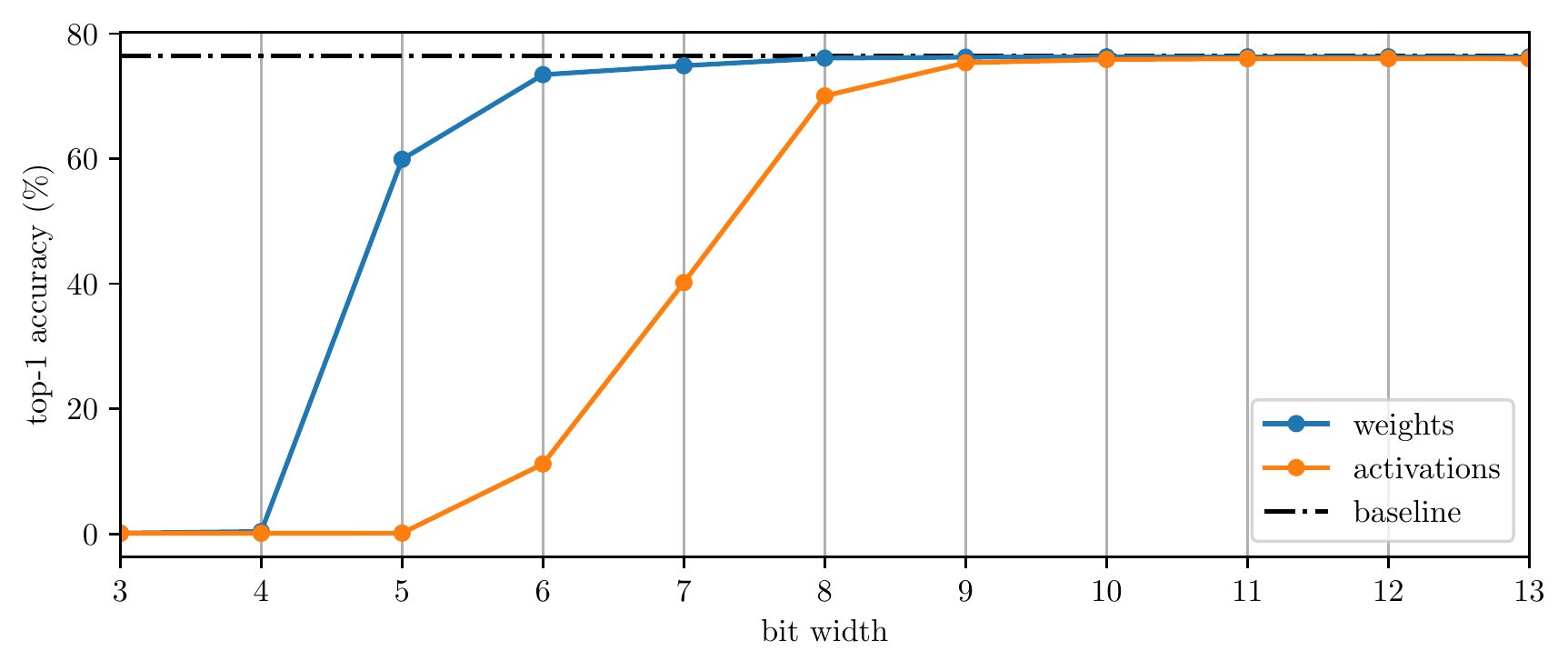}
\caption{Difference in sensitivity between activation and weight quantization of the EfficientNet-B0  model pre-trained on ImageNet. As a quantization scheme uniform quantization without re-training was used. Activations are more sensitive to quantization since model performance drops significantly faster. Going below 8 bit is challenging and often requires (quantization-aware) re-training of the model to compensate for the quantization error. Data originates from \cite{fanta4}.}
\label{weights_vs_acts}
\end{figure}

For neural network computing, the default precision used on general hardware like GPUs or CPUs is 32 bit floating-point (``single-precision''), which causes high computational costs, power consumption, arithmetic operation latency and memory requirements \cite{VS_survey}. Here, quantization techniques can also reduce the number of bits required to represent weight parameters and/or activations of the full-precision neural network, as they map the respective data values to a finite set of discrete quantization levels (clusters).
Providing $n$ such clusters allows to represent each data point in only $\log_2n$ bit. However, the continuous reduction of the number of clusters generally leads to an increasingly large error and degraded performances (see the EfficientNet-B0\footnote{\url{https://github.com/lukemelas/EfficientNet-PyTorch}, Apache License, Version 2.0 - Copyright (c) 2019 Luke Melas-Kyriazi} example in Figure~\ref{weights_vs_acts}). 

This trade-off is a well-known problem in information theory and is addressed by rate-distortion optimization, a concept in lossy data compression. It aims to determine the minimal number of bits per data symbol (bitrate) at which the reconstruction of the compressed data does not exceed a certain level of distortion. Applying this to the domain of neural network quantization, the objective is to minimize the bitrate of the weight parameters while keeping model degradation caused by quantization below a certain threshold, i.e., the predictive performance of the model should not be affected by reduced parameter precisions. 
In contrast to multimedia compression approaches, e.g., for audio or video coding, the compression of DNNs has unique challenges and opportunities. 
Foremost, the neural network parameters to be compressed are not perceived directly by a user, as e.g., for video data.
Therefore, the coding or compression error or distortion cannot be directly used as performance measure.
Instead, such accuracy measurement needs to be deducted from a subsequent inference step.
Then, current neural networks are highly over-parameterized \cite{denil2013predicting} which allows for high errors/differences between the full-precision and the quantized parameters (while still maintaining model performance).
Also, the various layer types and the location of a layer within the DNN have different impacts on the loss function, and thus different sensitivities to quantization.
\begin{figure}[t]
\centering
\includegraphics[width=0.75\textwidth]{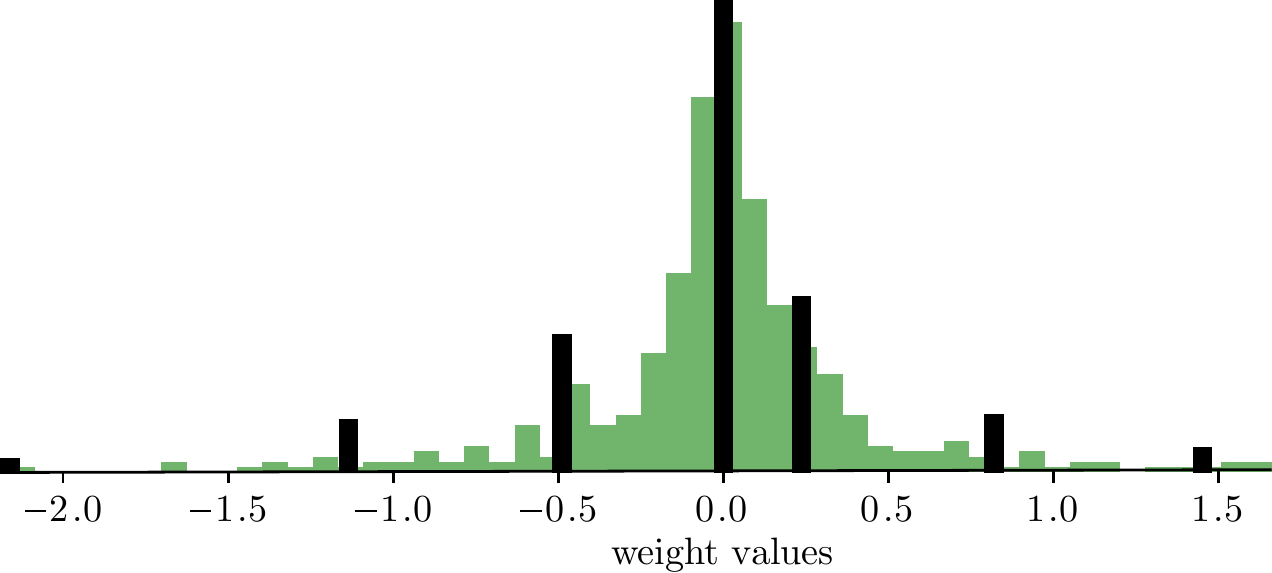}
\caption{Quantizing a neural network’s layer weights (binned weight distribution shown as green bars) to 7 discrete cluster centers (centroids).
The centroids (black bars) were generated by k-means clustering and the height of each bar represents the number of layer weights which are assigned to the respective centroid.} \label{clustering}
\end{figure}

Quantization can be further classified into uniform and non-uniform quantization.
The most intuitive way to initialize centroids is by arranging them equidistantly over the range of parameter values (uniform).
Other quantization schemes make use of non-uniform mapping functions, e.g., k-means clustering, which is determined by the distribution of weight values (see Figure \ref{clustering}).
As non-uniform quantization captures the underlying distribution of parameter values better, it may achieve less distortion compared to equidistantly arranged centroids.
However, non-uniform schemes are typically more difficult to deploy on hardware, e.g., they require a codebook (look-up table), whereas uniform quantization can be implemented using a single scaling factor (step size) which allows a very efficient hardware implementation with fixed-point integer logic.

\subsection{Entropy-Constrained Quantization}

As discussed in \cite{CER}, and experimentally shown in \cite{fanta4}, lowering the entropy of DNN weights provides benefits in terms of memory as well as computational complexity. 
The Entropy-Constrained Quantization (ECQ) algorithm is a clustering algorithm that also takes the entropy of the weight distributions into account. 
More precisely, the first-order entropy $H = -\sum_c P_c\log_2{P_c}$ is used,
where $P_c$ is the ratio of the number of parameter elements in the $c$-th cluster to the number of all parameter elements (i.e., the source distribution).
To recall, the entropy $H$ is the theoretical limit of the average number of bits required to represent any element of the distribution \cite{Shannon}. 

Thus, ECQ assigns weight values not only based on their distances to the centroids, but also based on the information content of the clusters. Similar to other rate-distortion-optimization methods, ECQ applies Lagrange optimization:

\begin{equation}\label{eq:assignment_cost_function}
\textbf{A}^{(l)} = \underset{c}{\text{argmin}} \ d(\textbf{W}^{(l)}, w_c^{(l)}) - \lambda^{(l)}\log_2(P_c^{(l)}).\\
\end{equation}

Per network layer $l$, the assignment matrix $\textbf{A}^{(l)}$ maps a centroid to each weight based on a minimization problem consisting of two terms: Given the full-precision weight matrix $\textbf{W}^{(l)}$ and the centroid values $w_c^{(l)}$, the first term in Equation~\eqref{eq:assignment_cost_function} measures the squared distance between all weight elements and the centroids, indexed by $c$.
The second term in Equation~\eqref{eq:assignment_cost_function} is weighted by the scalar Lagrange parameter $\lambda^{(l)}$ and describes the entropy constraint.
More precisely, the information content $I$ is considered, i.e., $I=-\log_2(P_c^{(l)})$, where the probability $P_c^{(l)}\in[0,1]$ defines how likely a weight element $w_{ij}^{(l)}\in\textbf{W}^{(l)}$ is going to be assigned to centroid $w_c^{(l)}$. 
Data elements with a high occurrence frequency, or a high probability, contain a low information content, and vice versa.
$P$ is calculated layer-wise as $P_c^{(l)} = N_{w_c}^{(l)} / N_{\textbf{W}}^{(l)}$, with $N_{w_c}^{(l)}$ being the number of full-precision weight elements assigned to the cluster with centroid value $w_c^{(l)}$ (based on the squared distance), and $N_{\textbf{W}}^{(l)}$ being the total number of parameters in $\textbf{W}^{(l)}$. Note that $\lambda^{(l)}$ is scaled with a factor based on the number of parameters a layer has in proportion to other layers in the network to mitigate the constraint for smaller layers.

The entropy regularization term motivates sparsity and low-bit weight quantization in order to achieve smaller coded neural network representations. Based on the specific neural network coding optimization, we developed ECQ. This algorithm is based on previous work in
Entropy-Constrained Trained Ternarization (EC2T) \cite{EC2T}.
EC2T trains sparse and ternary DNNs to state-of-the-art accuracies.

In our developed ECQ, we generalize the EC2T method, such that DNNs of variable bit width can be rendered. Also, ECQ does not train centroid values to facilitate integer arithmetic on general hardware. The proposed quantization-aware training algorithm includes the following steps:
\begin{enumerate}
    \item Quantize weight parameters by applying ECQ (but keep a copy of the full-precision weights).
    \item Apply Straight-Through Estimator (STE) \cite{STE}: 
    \begin{enumerate}
        \item Compute forward and backward pass through quantized model version.
        \item Update full-precision weights with scaled gradients obtained from quantized model.
    \end{enumerate}
\end{enumerate}
\section{Explainability-Driven Quantization} \label{sec:XQ}
Explainable AI techniques can be applied to find relevant features in input as well as latent space. 
Covering large sets of data, identification of relevant and functional model substructures is thus possible. 
Assuming over-parameterization of DNNs, the authors of~\cite{yeom2021pruning} exploit this for pruning (of irrelevant filters) to great effect. 
Their successful implementation shows the potential of applying XAI for the purpose of quantization as well, as sparsification is part of quantization,
e.g., by assigning weights to the zero-cluster.
Here, XAI opens up the possibility to go beyond regarding model weights as static quantities and to consider the interaction of the model with given (reference) data.
This work aims to combine the two orthogonal approaches of ECQ and XAI in order to further improve sparsity and efficiency of DNNs.
In the following, the LRP method is introduced, which can be applied to extract relevances of individual neurons, as well as weights.

\subsection{Layer-wise Relevance Propagation} \label{sec:lrp_intro}

Layer-wise Relevance Propagation (LRP) \cite{bach2015pixel} is an attribution method
based on the conservation of flows and proportional decomposition.
It explicitly is aligned to the layered structure of machine learning models.
Regarding a model with $n$ layers
\begin{equation} \label{eq:lrp_nn}
    f(x)=f_n \circ \dots \circ  f_1(x)~,
\end{equation}
LRP first calculates all activations during the forward pass starting with $f_1$ until the output layer $f_n$ is reached. Thereafter, the prediction score $f(x)$ of any chosen model output is redistributed layer-wise as an initial quantity of relevance $R_n$ back towards the input. 
During this backward pass, the redistribution process follows a conservation principle analogous to Kirchhoff’s laws in electrical circuits.
Specifically, all relevance that flows into a neuron is redistributed towards neurons of the layer below.
In the context of neural network predictors, the whole LRP procedure can be efficiently implemented as a forward–backward pass with modified gradient computation, as demonstrated in, e.g., \cite{samek2021explaining}.

Considering a layer's output neuron $j$, the distribution of its assigned relevance score $R_j$ towards its lower layer input neurons $i$ can be, in general, achieved by applying the basic decomposition rule
\begin{equation} \label{eq:lrp_basic_decomp}
    R_{i \leftarrow j} = \frac{z_{ij}}{z_j}R_j~,
\end{equation}
where $z_{ij}$ describes the contribution of neuron $i$ to the activation of neuron $j$ \cite{bach2015pixel,montavon2019layer}
and $z_j$ is the aggregation of the pre-activations $z_{ij}$ at output neuron $j$, i.e., $z_j = \sum_i z_{ij}$.
Here, the denominator enforces the conservation principle over all $i$ contributing to $j$, meaning $\sum_i R_{i \leftarrow j} =  R_j$. 
This is achieved by ensuring the decomposition of $R_j$ is in proportion to the relative flow of activations $z_{ij}/z_j$ in the forward pass.
The relevance of a neuron $i$ is then simply an aggregation of all incoming relevance quantities
\begin{equation} \label{eq:lrp_basic_aggregate}
    R_i = \sum_j R_{i \leftarrow j}~.
\end{equation}
Given the conservation of relevance in the decomposition step of Equation~\eqref{eq:lrp_basic_decomp}, this means that $\sum_i R_i = \sum_j R_j$ holds for consecutive neural network layers.
Next to component-wise non-linearities, linearly transforming layers (e.g., dense or convolutional) are by far the most common and basic building blocks of neural networks such as VGG-16 \cite{simonyan2014very} or ResNet \cite{he2016deep}.
While LRP treats the former via identity backward passes,
relevance decomposition formulas can be given for the latter explicitly in terms of weights $w_{ij}$ and input activations $a_i$.
Let the output of a linear neuron be given as $z_j = \sum_{i,0} z_{ij} = \sum_{i,0} a_i w_{ij}$ with bias ``weight'' $w_{0j}$ and respective activation $a_0=1$.
In accordance to Equation \eqref{eq:lrp_basic_decomp}, relevance is then propagated as
\begin{equation} \label{eq:lrp_linear}
     R_{i \leftarrow j}
     = \overbrace{\underbrace{a_i w_{ij}}_{z_{ij}}\frac{R_j}{z_j}}^{\text{explicit}}
     = a_i \overbrace{\underbrace{w_{ij}}_{\frac{\partial z_j}{\partial a_i}}\frac{R_j}{z_j}}^{\text{mod. grad.}}
     = w_{ij} \overbrace{\underbrace{a_i}_{\frac{\partial z_j}{\partial w_{ij}}} \frac{R_j}{z_j}}^{\text{mod. grad.}}~.
\end{equation}
Equation \eqref{eq:lrp_linear} exemplifies,
that the explicit computation of the backward directed relevances $R_{i \leftarrow j}$ in linear layers
can be replaced equivalently by a ``(modified gradient $\times$ input)'' approach.
Therefore, the activation $a_i$ or weight $w_{ij}$ can act as the input and target wrt. which the partial derivative regarding output $z_j$ is computed. The scaled relevance term $R_j / z_j$ takes the role of the upstream gradient to be propagated.

\begin{figure}[t]
    \centering
    \includegraphics[width=0.8\textwidth]{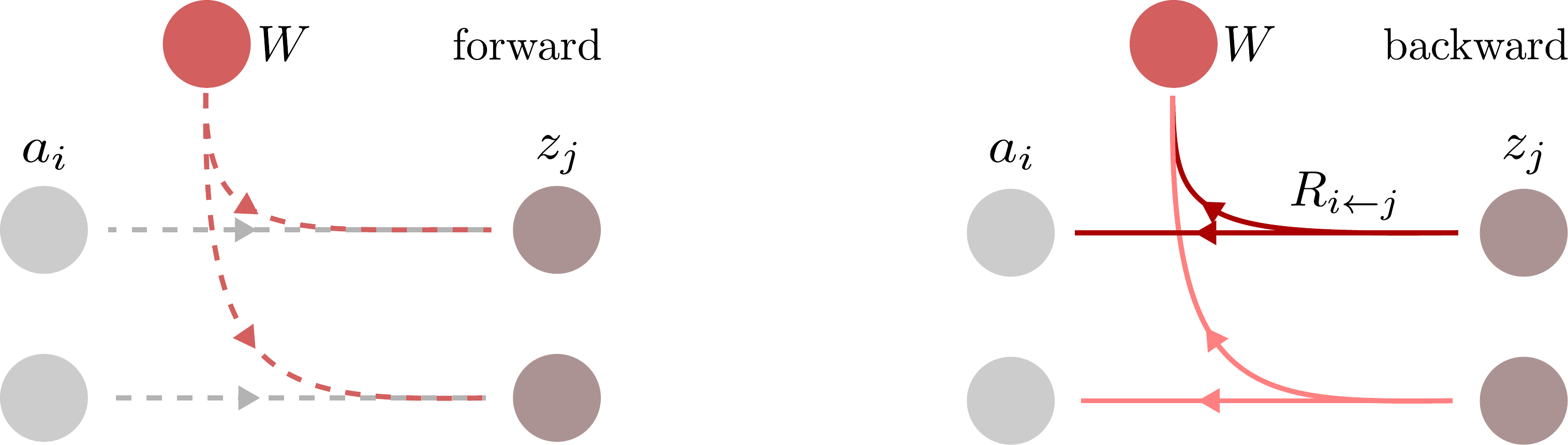}
    \caption{LRP can be utilized to calculate relevance scores for weight parameters $W$, which contribute to the activation of output neurons $z_j$ during the forward pass in interaction with data-dependent inputs $a_i$.
    In the backward pass, relevance messages $R_{i\leftarrow j}$ can be aggregated at neurons / input activations $a_i$, but also at weights $W$.}
    \label{fig:lrp_weight_distribution}
\end{figure} 

At this point, LRP offers the possibility to calculate relevances not only of neurons, but also of individual weights, depending on the aggregation strategy, as illustrated in Figure~\ref{fig:lrp_weight_distribution}.
This can be achieved by aggregating relevances at the corresponding (gradient) targets, i.e.,
plugging Equation~\eqref{eq:lrp_linear} into Equation~\eqref{eq:lrp_basic_aggregate}. 
For a dense layer, this yields
\begin{equation}
R_{w_{ij}} = R_{i \leftarrow j}
\label{eq:lrp_w_rule1}
\end{equation}
with an individual weight as the aggregation target contributing (exactly) once to an output.
A weight of a convolutional filter however is applied multiple times within a neural network layer.
Here, we introduce a variable $k$ signifying one such application context, e.g., one specific step in the application of a filter $w$ in a (strided) convolution, mapping the filter's inputs $i$ to an output $j$.
While the relevance decomposition formula within one such context $k$ does not change from Equation~\eqref{eq:lrp_basic_decomp},
we can uniquely identify its backwards distributed relevance messages as $R^k_{i \leftarrow j}$.
With that, the aggregation of relevance at the convolutional filter $w$ at a given layer is given with
\begin{equation}
R_{w_{ij}} = \sum_{k} R^k_{i \leftarrow j},
\label{eq:lrp_w_rule}
\end{equation}
where $k$ iterates over all applications of this filter weight. 

Note that in modern deep learning frameworks, derivatives wrt. activations or weights can be computed efficiently by leveraging the available automatic differentiation functionality (autograd) \cite{paszke2017automatic}. Specifying the gradient target, autograd then already merges the relevance decomposition and aggregation steps outlined above. Thus, computation of relevance scores for filter weights in convolutional layers is also appropriately supported,
for Equation~\eqref{eq:lrp_basic_decomp},
as well as any other relevance decomposition rule which can be formulated as a modified gradient backward pass, such as Equations~\eqref{eq:lrp_epsilon} and~\eqref{eq:lrp_alpha}.
The ability to compute the relevance of individual weights is a critical ingredient for the eXplainability-driven Entropy-Constrained Quantization strategy introduced in Section~\ref{subsec:LRPquant}.

In the following, we will briefly introduce further LRP decomposition rules used throughout our study. In order to increase numerical stability of the basic decomposition rule in Equation~\eqref{eq:lrp_basic_decomp}, the LRP $\varepsilon$-rule introduces a small term $\varepsilon$ in the denominator:
\begin{equation} \label{eq:lrp_epsilon}
    R_{i\leftarrow j} = \frac{z_{ij}}{z_j + \varepsilon \cdot \text{sign}(z_j)} R_j~.
\end{equation}
The term $\varepsilon$ absorbs relevance for weak or contradictory contributions to the activation of neuron $j$.
Note here, in order to avoid divisions by zero, the $\text{sign}(z)$ function is defined to return 1 if $z \geq 0$ and -1 otherwise.
In the case of a deep rectifier network, it can be shown \cite{ancona2019gradient} that the application of this rule to the whole neural network results in an explanation that is similar to (simple) (gradient $\times$ input) \cite{shrikumar2016not}.
A common problem within deep neural networks is,
that the gradient becomes increasingly noisy with network depth \cite{samek2021explaining}, partly a result from gradient shattering~\cite{balduzzi2017shattered}.
The $\varepsilon$ parameter is able to suppress the influence of that noise given sufficient magnitude.
With the aim of achieving robust decompositions,
several purposed rules next to Equations~\eqref{eq:lrp_basic_decomp} and \eqref{eq:lrp_epsilon} have been proposed in literature (see \cite{montavon2019layer} for an overview).

One particular rule choice, which reduces the problem of gradient shattering and which has been shown to work well in practice, is the $\alpha \beta$-rule~\cite{bach2015pixel,montavon2018methods}
\begin{equation} \label{eq:lrp_alpha}
    R_{i\leftarrow j} = \left( \alpha \frac{\left(z_{ij}\right)^+}{\left(z_j\right)^+} - \beta \frac{\left(z_{ij}\right)^-}{\left(z_j\right)^-} \right) R_j~,
\end{equation}
where $(\cdot)^+$ and $(\cdot)^-$ denote the positive and negative parts of the variables $z_{ij}$ and $z_j$, respectively.
Further, the parameters $\alpha$ and $\beta$ are chosen subject to the constraints $\alpha - \beta = 1$ and $\beta \geq 0$ (i.e.,  $\alpha \geq 1$) in order to  propagate relevance conservatively throughout the network.
Setting $\alpha=1$, the relevance flow is computed only with respect to the positive contributions $\left(z_{ij}\right)^+$ in the forward pass.
When alternatively parameterizing with, e.g., $\alpha = 2$ and $\beta = 1$, which is a common choice in literature,
negative contributions are included as well, while favoring positive contributions. 

Recent works recommend a composite strategy of decomposition rule assignments mapping multiple rules purposedly to different parts of the network~\cite{montavon2019layer,kohlbrenner2020towards}. 
This leads to an increased quality of relevance attributions for the intention of explaining prediction outcomes.
In the following, 
a composite strategy consisting of the $\varepsilon$-rule for dense layers and the $\alpha\beta$-rule with $\beta=1$ for convolutional layers is used.
Regarding LRP-based pruning,
Yeom et al.~\cite{yeom2021pruning} utilize the $\alpha \beta$-rule \eqref{eq:lrp_alpha} with $\beta=0$ for convolutional as well as dense layers. 
However, using $\beta=0$, subparts of the network that contributed solely negatively, might receive no relevance. 
In our case of quantization, all individual weights have to be considered. 
Thus, the $\alpha\beta$-rule with $\beta=1$ is used for convolutional layers, 
because it also includes negative contributions in the relevance distribution process and reduces gradient shattering. 
The LRP implementation is based on the software package Zennit~\cite{anders2021software}, 
which offers a flexible integration of composite strategies and readily enables extensions required for the computation of relevance scores for weights.

\subsection{eXplainability-driven Entropy-Constrained Quantization}\label{subsec:LRPquant}

For our novel eXplainability-driven Entropy-Constrained Quantization (ECQ$^{\text{x}}$), we modify the ECQ assignment function to optimally re-assign the weight clustering based on LRP relevances in order to achieve higher performance measures and compression efficiency.
The rationale behind using LRP to optimize the ECQ quantization algorithm is two-fold: 

\paragraph{Assignment correction:} 
In the quantization process, the entropy regularization term encourages weight assignments to more populated clusters in order to minimize the overall entropy. Since weights are usually normally distributed around zero, the entropy term also strongly encourages sparsity. In practice, this quantization scheme works well rendering sparse and low-bit neural networks for various machine learning tasks and network architectures \cite{fanta4,EC2T,ECT}. 

From a scientific point of view, however, one might wonder why the shift of numerous weights from their nearest-neighbor clusters to a more distant cluster does not lead to greater model degradation, especially when assigned to zero. The quantization-aware re-training and fine-tuning can, up to a certain extent, compensate for this shift. Here, the LRP-generated relevances show potential to further improve quantization in two ways: 1) by re-adding ``highly relevant'' weights (i.e., preventing their assignment to zero if they have a high relevance), and 2) by assigning additional, ``irrelevant'' weights to zero (i.e., preventing their distance- and entropy-based assignment to a non-zero centroid).

\begin{figure}[t!]
\centering
\includegraphics[width=0.495\textwidth]{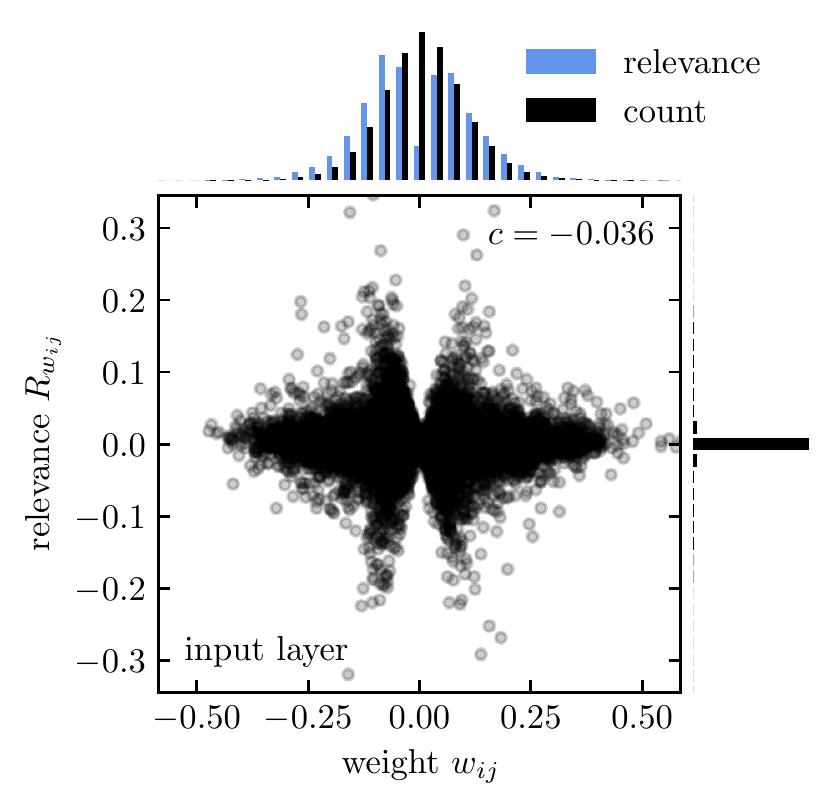}
\includegraphics[width=0.495\textwidth]{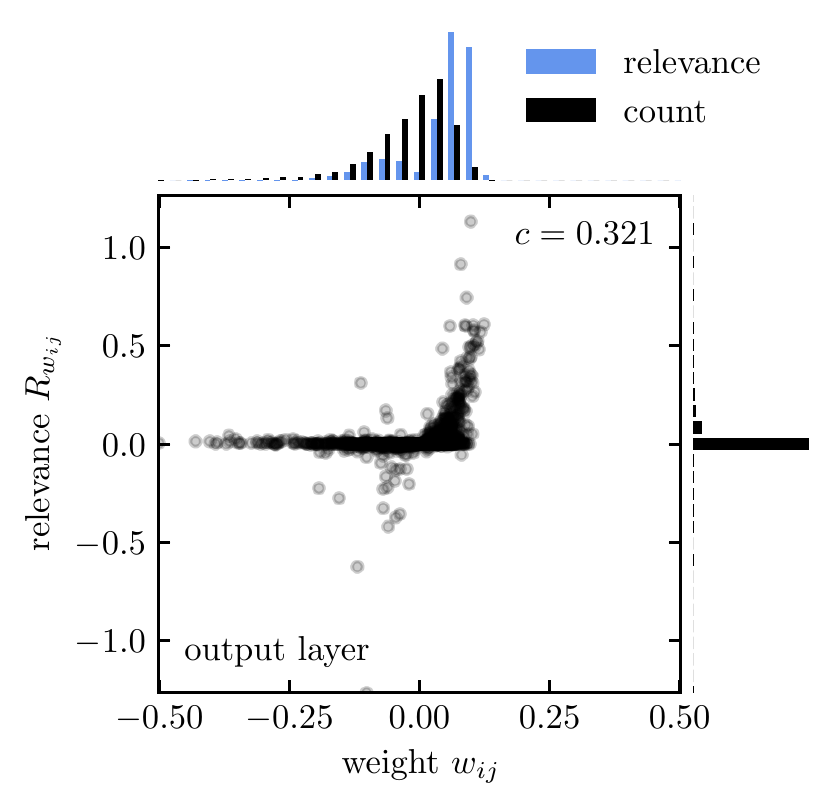}
\caption{Weight relevance $R_{w_{ij}}$ vs. weight value $w_{ij}$ for the input layer (left) and output layer (right) of the full-precision MLP$\_$GSC model (introduced in Section \ref{subsec:exp_setup}).
The black histograms to the top and right of each panel display the distributions of weights (top) and relevances (right).
The blue histograms further show the amount of relevance (blue) of each weight histogram bin.
All relevances are collected over the validation set with equally weighted samples (i.e., by choosing $R_n = 1$).
The value $c$ measures the Pearsson correlation coefficient between weights and relevances.
}
\label{fig:rel_magn} 
\end{figure}
We evaluated the discrepancy between weight relevance and magnitude in a correlation analysis depicted in Figure \ref{fig:rel_magn}. Here, all weight values $w_{ij}$ are plotted against their associated relevance $R_{w_{ij}}$ for the input layer (left) and output layer (right) of the full-precision model MLP$\_$GSC (which will be introduced in Section \ref{subsec:exp_setup}).
In addition, histograms of both parameters are shown above and to the right of each relevance-weight-chart in Figure \ref{fig:rel_magn} to better visualize the correlation between $w_{ij}$ and $R_{w_{ij}}$.
In particular, a weight of high magnitude is not necessarily also a relevant weight. And in contrast, there are also weights of small or medium magnitude that have a high relevance and thus should not be omitted in the quantization process. 
This phenomenon is especially true for layers closer to the input. 
The outcome of this analysis strongly motivates the use of LRP relevances for the weight assignment correction process of low-bit and sparse ECQ$^{\text{x}}$.

\paragraph{Regularizing effect for training:}
Since the previously described re-adding (which is also referred to as ``regrowth'' in literature) and removing of weights due to LRP depends on the propagated input data, weight relevances can change from data batch to data batch. In our quantization-aware training, we apply the STE, and thus the re-assignment of weights, after each forward-backward pass. 

The regularizing effect which occurs due to dynamic re-adding and removing weights is probably related to the generalization effect which random Dropout \cite{srivastava2014dropout} has on neural networks. However, as elaborated in the extensive survey by Hoefler et al. \cite{hoefler2021sparsity}, in terms of dynamic sparsification, re-adding (``drop in'') the best weights is as crucial as removing (``drop out'') the right ones. Instead of randomly dropping weights, the work in \cite{dai2019nest} shows that re-adding weights based on largest gradients is  related to Hebbian learning and biologically more plausible. LRP relevances go beyond the gradient criterion, which is why we consider it a suitable candidate.\\

In order to embed LRP relevances in the assignment function (\ref{eq:assignment_cost_function}), 
we update the cost for the zero centroid ($c=0$) by extending it as

\begin{equation}\label{eq:assignment_cost_function_lrp}
\rho~\textbf{R}_{W^{(l)}} \cdot \left( d(\textbf{W}^{(l)}, w_{c=0}^{(l)}) - \lambda^{(l)}~\log_2(P_{c=0}^{(l)}) \right)
\end{equation}
with relevance matrix $\textbf{R}_{W^{(l)}}$ containing all weight relevances $R_{w_{ij}}$ of layer $l$ with row/input index $i$ and column/output index $j$, as specified in Equation~\eqref{eq:lrp_w_rule}.
The relevance-dependent assignment matrix $\textbf{A}_{\text{x}}^{(l)}$ is thus described by:
\begin{equation}\label{eq:full_assignment_cost_function_lrp}
\textbf{A}_{\text{x}}^{(l)}(\textbf{W}^{(l)}) = \underset{c}{\text{argmin}}
\begin{cases}
\rho~\textbf{R}_{W^{(l)}} \cdot \left( d(\textbf{W}^{(l)}, w_{c=0}^{(l)}) - \lambda^{(l)}~\log_2(P_{c=0}^{(l)}) \right) & \text{, if } c = 0\\
\\
d(\textbf{W}^{(l)}, w_c^{(l)}) - \lambda^{(l)}~\log_2(P_c^{(l)}) & \text{, if } c \neq 0
\end{cases}
\end{equation}

where $\rho$ is a normalizing scaling factor, which also takes relevances of the previous data batches into account (momentum).
The term $\rho~R_W$ increases the assignment cost of the zero cluster for relevant weights and decreases it for irrelevant weights.

\begin{figure}[!t]
    \centering
    \includegraphics[width=1.0\textwidth]{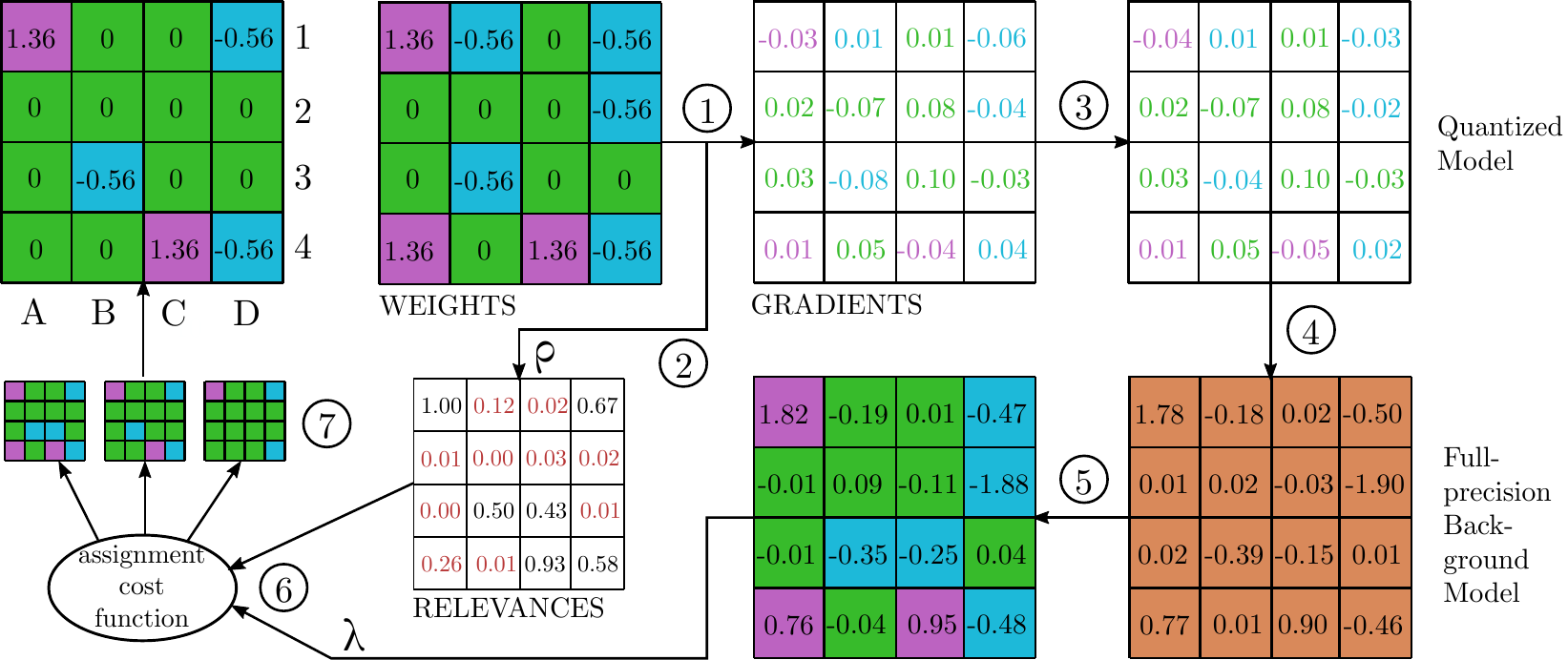}
    \caption{Exemplary ECQ$^{\text{x}}$ weight update. For simplicity, 3 centroids are used (i.e., symmetric 2 bit case). The process involves the following steps: 1) Derive gradients and LRP relevances from forward-backward pass. 2) LRP relevance scaling. 3) Gradients scaling. 4) Gradient attachment to full precision background model. 5) Background model update and nearest-neighbor clustering. 6) Computing of the assignment cost for each weight using the $\lambda$-scaled information content of clusters and the $\rho$-scaled relevances. Assign each weight by minimizing the cost. 7) Choosing an appropriate candidate (of various $\lambda$ and $\rho$ settings).}
    \label{fig:ECTx}
\end{figure}

Figure~\ref{fig:ECTx} shows an example of one ECQ$^{\text{x}}$ iteration that includes the following steps: 
1) ECQ$^{\text{x}}$ computes a forward-backward pass through the quantized model, deriving its weight gradients. LRP relevances $\textbf{R}_W$ are computed by redistributing modified gradients according to Equation~\eqref{eq:lrp_w_rule}.
2) LRP relevances are then scaled by a normalizing scaling factor $\rho$,
and  3) weight gradients are scaled by multiplying the non-zero centroid values (e.g., the upper left gradient of $-0.03$ is multiplied by the centroid value $1.36$).
4) The scaled gradients are then applied to the full-precision (FP) background model which is a copy of the initial unquantized neural network and is used only for weight assignment, i.e. it is updated with the scaled gradients of the quantized network but does not perform inference itself,
5) The FP model is updated using the ADAM optimizer \cite{ADAM}. Then, weights are assigned to their nearest-neighbor cluster centroids.
6) Finally, the assignment $\textbf{A}_{\text{x}}$ cost for each weight to each centroid  is calculated using the $\lambda$-scaled information content of clusters (i.e., $I_{- \text{ (blue)}}\approx 1.7$, $I_{0 \text{ (green)}}=1.0$ and $I_{+ \text{ (purple)}}\approx 2.4$ in this example) and $\rho$-scaled relevances.
Here, relevances above the exemplary threshold (i.e., mean $\bar{\textbf{R}}_W\approx 0.3$) increase the cost for the zero cluster assignment, while relevances below (highlighted in red) decrease it.
Each weight is assigned such that the cost function is minimized according to~Equation~\eqref{eq:full_assignment_cost_function_lrp}.
7) Depending on the intensity of the entropy and relevance constraints (controlled by $\lambda$ and $\rho$),
different assignment candidates can be rendered to fit a specific deep learning task.
In the example shown in Figure~\ref{fig:ECTx}, an exemplary candidate grid was selected, which is depicted at the top left of the Figure.
The weight at grid coordinate $D2$, for example, was assigned to the zero cluster due to its irrelevance and
the weight at $C3$ due to the entropy constraint.

In the case of dense or convolutional layers, LRP relevances can be computed efficiently using the autograd functionality, as mentioned in Section~\ref{sec:lrp_intro}.
For a classification task, it is sensible to use the target class score as a starting point for the LRP backward pass. 
This way, the relevance of a neuron or weight describes its contribution to the target class prediction. Since the output is propagated throughout the network, all relevance is proportional to the output score.
Consequently, relevances of each sample in a training batch are, in general, weighted differently according to their respective model output, or prediction confidence. 
However, with the aim of suppressing relevances for inaccurate predictions, 
it is sensible to weigh samples according to the model output, 
because a low output score usually corresponds to an unconfident decision of the model.

After the relevance calculation of a whole data batch, the relevance scores $\textbf{R}_{W^{(l)}}$ are transformed to their absolute value and normalized, such that $\textbf{R}_{W^{(l)}} \in[0, 1]$.
Even though negative contributions work against an output, 
they might still be relevant to the network functionality, and their influence is thus considered instead of omitted. 
On one hand, they can lead to positive contributions for other classes. 
On the other, they can be relevant to balancing neuron activations throughout the network. 

The relevance matrices $\textbf{R}_{W^{(l)}}$ resulting from LRP are usually sparse, as can be seen in the weight histograms of Figure \ref{fig:rel_magn}.
In order to control the effect of LRP in the assignment function, the relevances are exponentially transformed by $\beta$, applying a similar effect as for gamma correction in image processing:
\begin{equation*}
    \textbf{R}_{W^{(l)}}' = \left(\textbf{R}_{W^{(l)}}\right)^\beta
\end{equation*}
with $\beta \in [0, 1]$. 
Here, the parameter $\beta$ is initially chosen such that the mean relevance $\hat R_{W^{(l)}}$ does not change the assignment, e.g.,
$\rho  \left(\hat R_{W^{(l)}} \right)^\beta = 1$ or 
$\beta = -\frac{\ln{\rho}}{\ln{\hat R_{W^{(l)}}}}$. In order to further control the sparsity of a layer, the target sparsity $p$ is introduced. If the assignment increases a layer's sparsity by more than the target sparsity $p$, parameter $\beta$ is accordingly minimized. Thus, in ECQ$^{\text{x}}$, LRP relevances are directly included in the assignment function and their effect can be controlled by parameter $p$. An experimental validation of the developed ECQ$^{\text{x}}$ method, including state-of-the-art comparison and parameter variation tests, is given in the following section.

\section{Experiments} \label{sec:exp}

In the experiments, we evaluate our novel quantization method ECQ$^{\text{x}}$ using two widely used neural network architectures, namely a convolutional neural network (CNN) and a multilayer perceptron (MLP).
More precisely, we deploy VGG16 for the task of small-scale image classification (CIFAR-10), ResNet18 for the Pascal Visual Object Classes Challenge (Pascal VOC) and an MLP with 5 hidden layers and ReLU non-linearities solving the task of keyword spotting in audio data (Google Speech Commands).
 
In the first subsection, the experimental setup and test conditions are described, while the results are shown and discussed in the second subsection. In particular, results for ECQ$^{\text{x}}$ hyperparameter variation are shown, followed by a comparison against classical ECQ and results for bit width variation. Finally, overall results for ECQ$^{\text{x}}$ for different accuracy and compression measurements are shown and discussed. 

\subsection{Experimental Setup}\label{subsec:exp_setup}
All experiments were conducted using the PyTorch deep learning framework, version 1.7.1 with torchvision 0.8.2 and torchaudio 0.7.2 extensions. As a hardware platform we used Tesla V100 GPUs with CUDA version 10.2. 
The quantization-aware training of ECQ$^{\text{x}}$ was executed for 20 epochs in all experiments. As an optimizer we used ADAM with an initial learning rate of 0.0001.
In the scope of the training procedure, we consider \emph{all} convolutional and fully-connected layers of the neural networks for quantization, including the input and output layers.
Note that numerous approaches in related works keep the input and/or output layers in full-precision (32 bit float), which may compensate for the model degradation caused by quantization, but is usually difficult to bring into application and incurs significant overhead in terms of energy consumption. 

\subsubsection{Google Speech Commands}
The Google Speech Commands (GSC \cite{warden_speech_2018}) dataset consists of 105,829 utterances of 35 words recorded from 2,618 speakers. The standard is to discriminate ten words ``Yes'', ``No'', ``Up'', ``Down'', ``Left'', ``Right'', ``On'', ``Off'', ``Stop'', and ``Go'', and adding two additional labels, one for ``Unknown Words'', and another for ``Silence'' (no speech detected).
Following the official Tensorflow example code for training\footnote{\url{https://github.com/tensorflow/tensorflow/tree/master/tensorflow/examples/speech\_commands}}, we implemented the corresponding data augmentation with PyTorch's torchaudio package. It includes randomly adding background noise with a probability of 80$\%$ and time shifting the audio [$-100, 100$] ms with a probability of 50$\%$. To generate features, the audio is  transformed to  MFCC fingerprints (Mel Frequency Cepstral Coefficients). We use 15 bins and a window length of 2000 ms. 
To solve GSC, we deploy a MLP (which we name \textit{MLP\_GSC} in the following) consisting of an input layer, five hidden layers and an output layer featuring 512, 512, 256, 256, 128, 128 and 12 output features, respectively. The MLP\_GSC was pre-trained for 100 epochs using stochastic gradient descent (SGD) optimization with a momentum of 0.9, an initial learning rate of 0.01 and a cosine annealing learning rate schedule.

\subsubsection{CIFAR-10}
The CIFAR-10 \cite{krizhevsky_learning_2009} dataset consists of natural images with a resolution of $32\times 32$ pixels. It contains 10 classes, with 6,000 images per class. Data is split to 50,000 training and 10,000 test images.
We use standard data pre-processing, i.e., normalization, random horizontal flipping and cropping. To solve the task, we deploy a VGG16 from the torchvision model zoo\footnote{\label{torchvision}\url{https://pytorch.org/vision/stable/models.html}}. The VGG16 classifier is adapted from 1,000 ImageNet classes to ten CIFAR classes by replacing its three fully-connected layers (with dimensions [25,088, 4,096], [4,096, 4,096], [4,096, 1,000]) by two ([512, 512], [512, 10]), as a consequence of CIFAR's smaller image size. We also implemented a VGG16 supporting batch normalization (``BatchNorm'' in the following), i.e. VGG16\_bn from torchvision. The VGGs were transfer-learned for 60 epochs using ADAM optimization and an initial learning rate of 0.0005.

\subsubsection{Pascal VOC}
The Pascal Visual Object Classes Challenge 2012 (VOC2012) \cite{pascal-voc-2012} provides 11,540 images associated with 20 classes.
The dataset has been split into 80$\%$ for training/validation and 20$\%$ for testing. We applied normalization, random horizontal flipping and center cropping to $224\times 224$ pixels. As a neural network architecture, the pre-trained ResNet18 from the torchvision model zoo\footnotemark[\value{footnote}] was deployed. Its classifier was adapted to predict 20 instead of 1,000 classes and the model was transfer-learned for 30 epochs using ADAM optimization with an initial learning rate of 0.0001.

\subsection{ ECQ$^{\text{x}}$ Results}
In this subsection, we compare ECQ$^{\text{x}}$ to state-of-the-art ECQ quantization, analysing accuracy preservation vs. sparsity increase. Furthermore, we investigate ECQ$^{\text{x}}$ compressibility, behavior on BatchNorm layers, and an appropriate choice of hyperparameters.

\subsubsection{ECQ$^{\text{x}}$ Hyperparameter Variation}

\begin{figure}[t]
\centering
\includegraphics[width=1.0\textwidth]{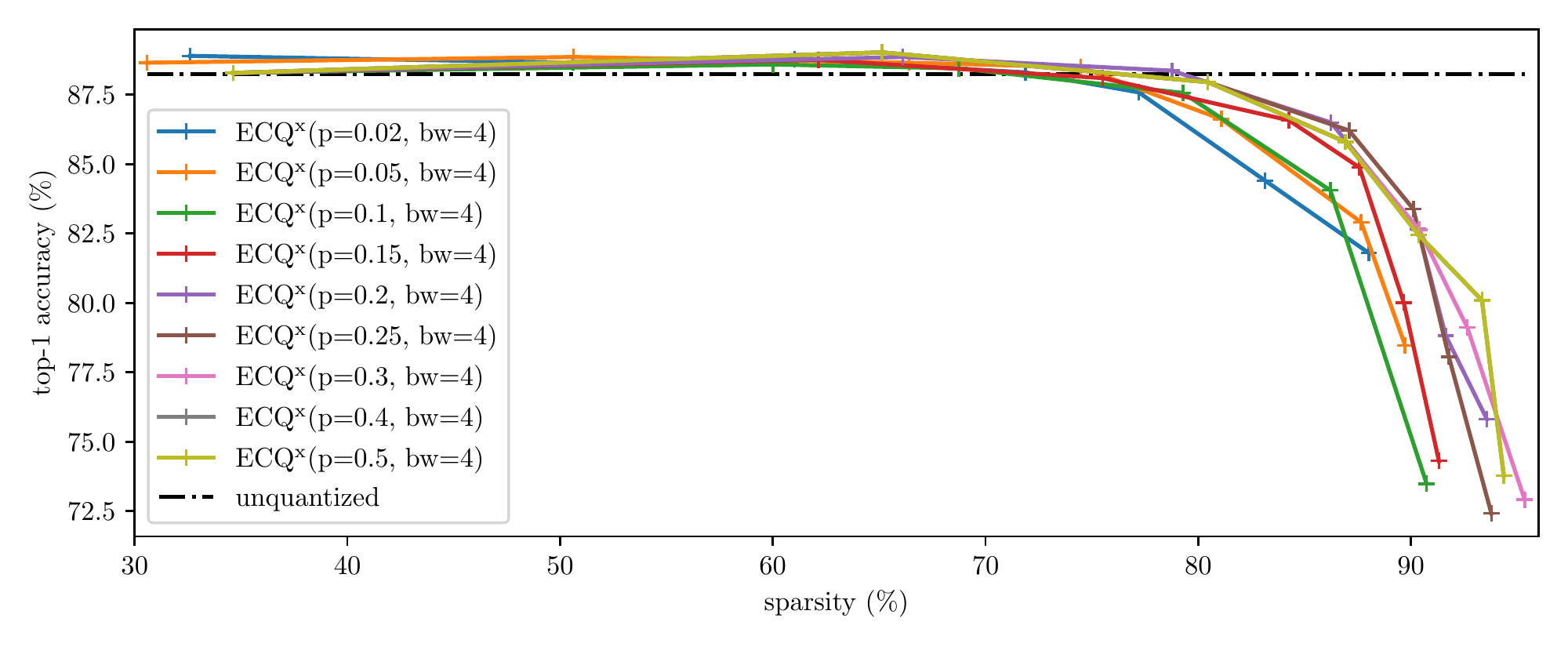}
\caption{Hyperparameter $p$ controls the LRP-introduced sparsity.} \label{fig:p_hyperparam}
\end{figure}

In ECQ$^{\text{x}}$, two important hyperparameters, $\lambda$ and $p$, influence the performance and thus are optimized for the comparative experiments described below. $\lambda$ increases the intensity of the entropy constraint and thus distributes the working points of each trial over a range of sparsities (see Figure \ref{fig:p_hyperparam}). The $p$ hyperparameter defines  an upper bound for the per-layer percentage of zero values, allowing a maximum amount of $p$ additional sparsity, on top of the $\lambda$-introduced sparsity. It thus implicitly controls the intensity of the LRP constraint.
Figure \ref{fig:p_hyperparam} shows results using several $p$ values for the 4 bit ($bw=4$) quantization of the MLP\_GSC model. Note, that the variation of bit width $bw$ is discussed below the comparative results. For smaller $p$, less sparse models are rendered with higher top-1 accuracies in the low-sparsity regime (e.g., $p=0.02$ or $p=0.05$ between 30-50$\%$ total network sparsity). In the regime of higher sparsity, larger values of $p$ show a better sparsity-accuracy trade-off. Note, that larger $p$ do not only set more weights to zero but also re-add relevant weights (regrowth). 
For $p=0.4$ and $p=0.5$, both lines are congruent since no layer is achieving more than $40\%$ additional LRP-introduced sparsity with the initial $\beta$ value (cf. Section \ref{subsec:LRPquant}). 

\subsubsection{ECQ$^{\text{x}}$ vs. ECQ Analysis}
\begin{figure}[t!]
\centering
\includegraphics[width=1.0\textwidth]{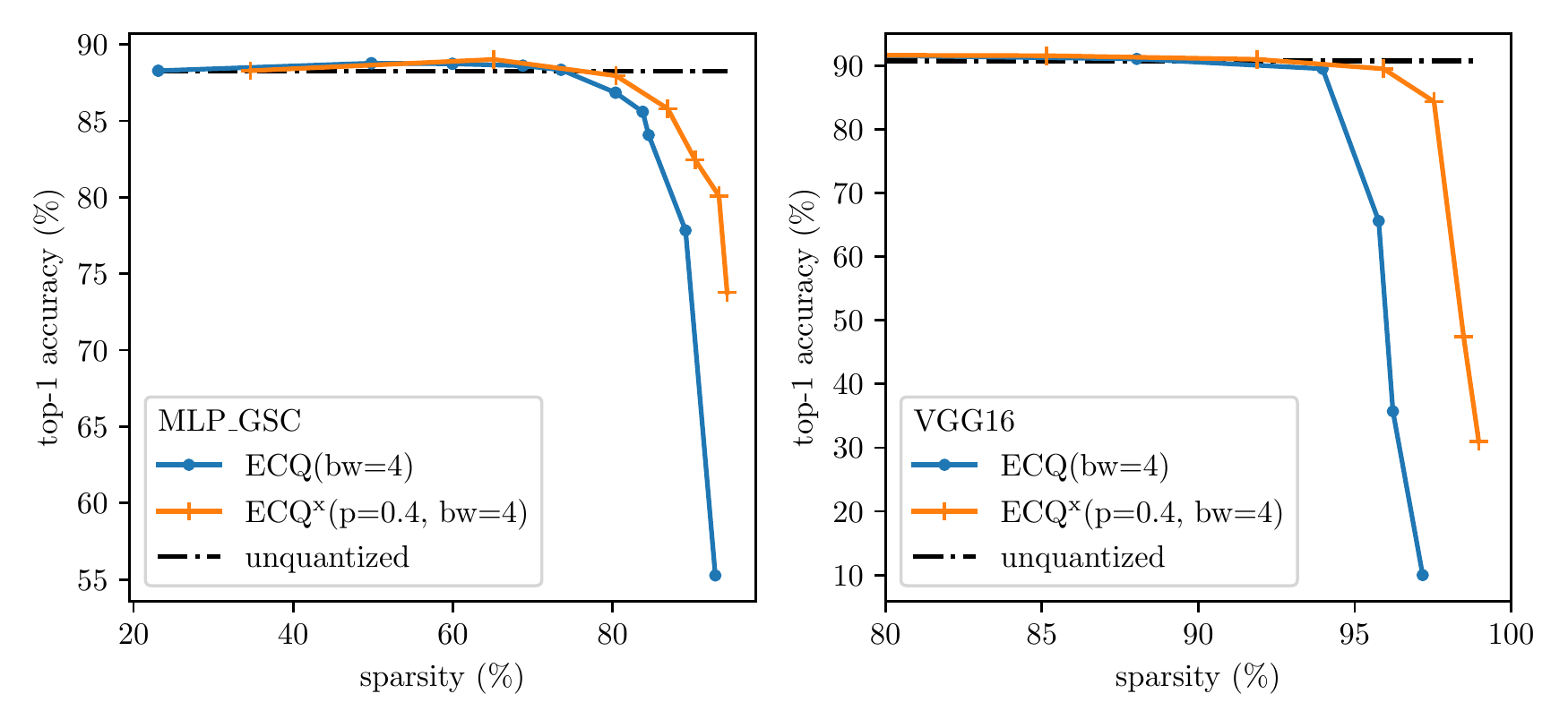}
\caption{Resulting model performances, when applying ECQ vs. ECQ$^{\text{x}}$ 4 bit quantization on MLP$\_$GSC (left) and VGG16 (right). Each point corresponds to a model rendered with a specific $\lambda$ which is a regulator for the entropy constraint and thus incrementally enhances sparsity. Abbreviations in the legend labels refer to bit width ($bw$) and target sparsity ($p$), which is defined in \ref{subsec:LRPquant}.} \label{fig:ecq_vs_ecqx}
\end{figure}
\begin{figure}[t!]
\centering
\includegraphics[width=1.0\textwidth]{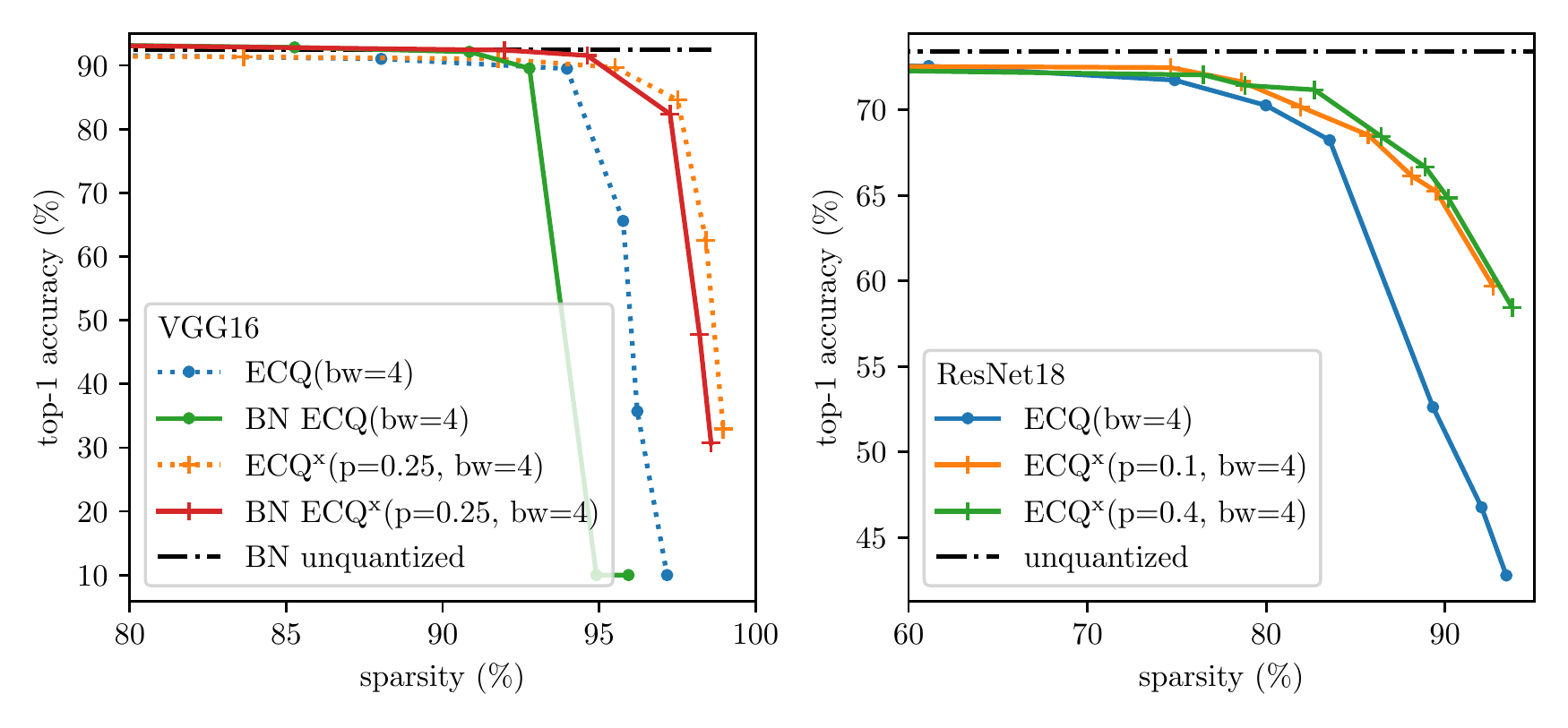}
\caption{Resulting model performances, when applying ECQ vs. ECQ$^{\text{x}}$ 4 bit quantization on VGG16, VGG16 with BatchNorm (BN) modules (left) and ResNet18 (right).} \label{fig:ecqx_BN}
\end{figure}
As shown in Figure \ref{fig:ecq_vs_ecqx}, the LRP-driven ECQ$^{\text{x}}$ approach renders models with higher performance and simultaneously higher efficiency. In this comparison, efficiency is determined in terms of sparsity, which can be exploited to compress the model more or to skip arithmetic operations with zero values. Both methods achieve a quantization to 4 bit integer without any performance degradation of the model. Performance is even slightly increased due to quantization when compared to the unquantized baseline. In the regime of high sparsity, model accuracy of the previous state-of-the-art (ECQ) drops significantly faster compared to the LRP-adjusted quantization scheme.

Regarding the handling of BatchNorm modules for LRP, 
it is proposed in literature to merge the BatchNorm layer parameters with the preceding linear layer \cite{guillemot2020breaking} into a single linear transformation. 
This canonization process is sensible, 
because it reduces the number of computational steps in the backward pass while maintaining functional equivalence between the original and the canonized model in the forward pass.

It has been further shown, that network canonization can increase explanation quality~\cite{guillemot2020breaking}.
With the aim of computing weight relevance scores for a BatchNorm layer's adjacent linear layer in its original (trainable) state,
keeping the layers separate is more favorable than merging.
Therefore, the $\alpha \beta$-rule with $\beta=1$ is also applied to BatchNorm layers.
The quantization results of the VGG architecture with BatchNorm modules and ResNet18 are shown in Figure~\ref{fig:ecqx_BN}.

In order to capture the computational overhead of LRP in terms of additional training time, we compared the average training times of the different model architectures per epoch. Relevance-dependent quantization (ECQ$^{\text{x}}$) requires approximately $1.2\times$, $2.4\times$, and $3.2\times$ more processing time than baseline quantization (ECQ) for the MLP\_GSC, VGG16, and ResNet18 architectures, respectively. This extra effort can be explained with the additional forward-backward passes performed in Zennit for LRP computation. More concretely, using Zennit as a plug-in XAI module, it computes one additional forward pass layer-wise and redistributes the relevances to the preceding layers according to the decomposition and aggregation rules specified in Section \ref{sec:lrp_intro}. For redistribution, Zennit computes one additional backward pass for $\varepsilon$-rule associated layers and two additional backward passes for $\alpha \beta$-rule associated layers in order to derive positive $\alpha$ and negative $\beta$ relevance contributions. To recap, in the applied composite strategy, the $\varepsilon$-rule is used for dense layers and the $\alpha \beta$-rule for convolutional layers and BatchNorm parameters, which results in the extra computational cost for VGG16 and ResNet18 compared to MLP\_GSC, which consists solely of dense layers. In addition, aggregation of relevances for convolutional filters is not required for dense layers. Note that the above mentioned values for additional computational overhead of ECQ$^{\text{x}}$ due to relevance computation can be interpreted as an upper-bound and that there are options to minimize the effort, e.g. by 1) not considering relevances for cluster assignments in each training iteration, 2) leveraging pre-computed outputs or even gradients from the quantized base model instead of separately computing forward-backward passes with a model copy in the Zennit module. Whereas 1) corresponds to a change in the quantization setup, 2) requires parallelization optimizations of the software framework.

\subsubsection{Bit Width Variation}

\begin{figure}[!t]
\centering
\includegraphics[width=1.0\textwidth]{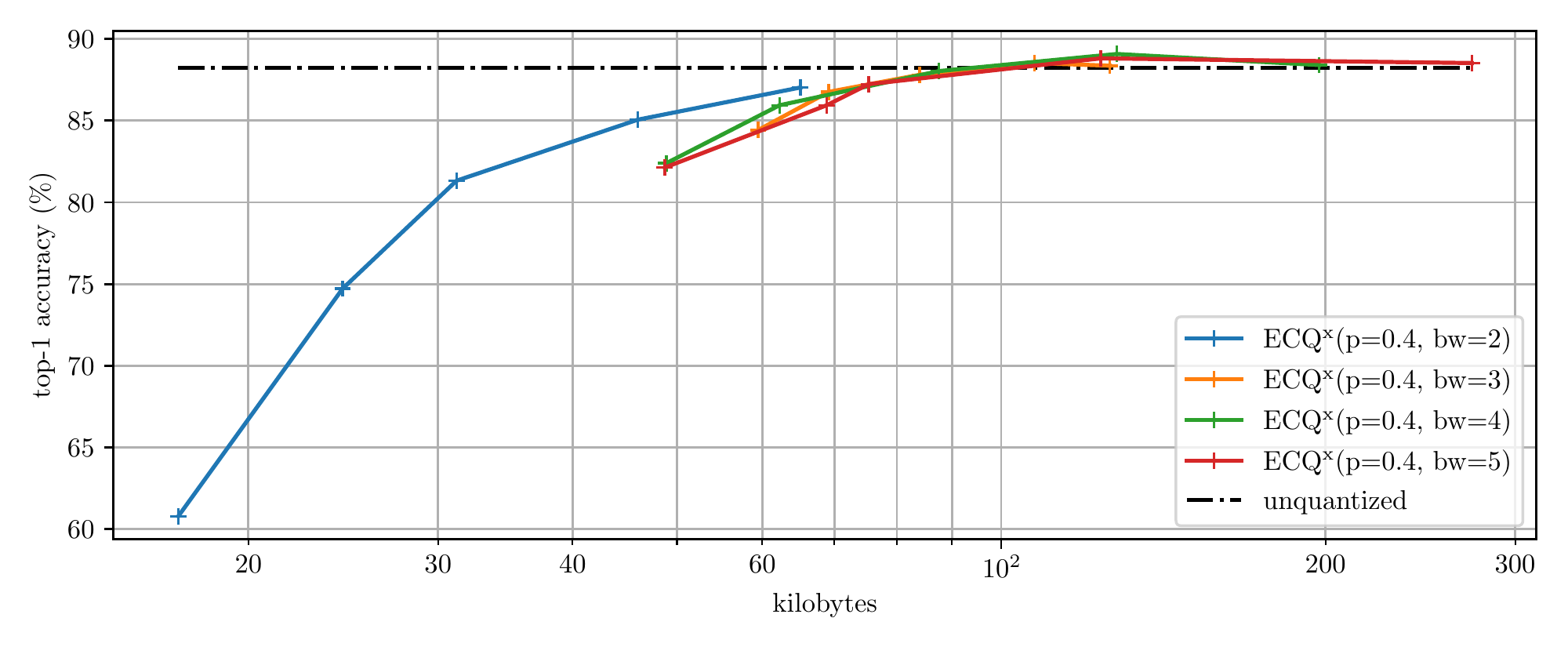}
\caption{Resulting MLP$\_$GSC model performances vs. memory footprint, when applying ECQ$^{\text{x}}$ with 2 bit to 5 bit quantization.} \label{fig:bitwidth_mlp}
\end{figure}

\begin{figure}[!t]
\centering
\includegraphics[width=1.0\textwidth]{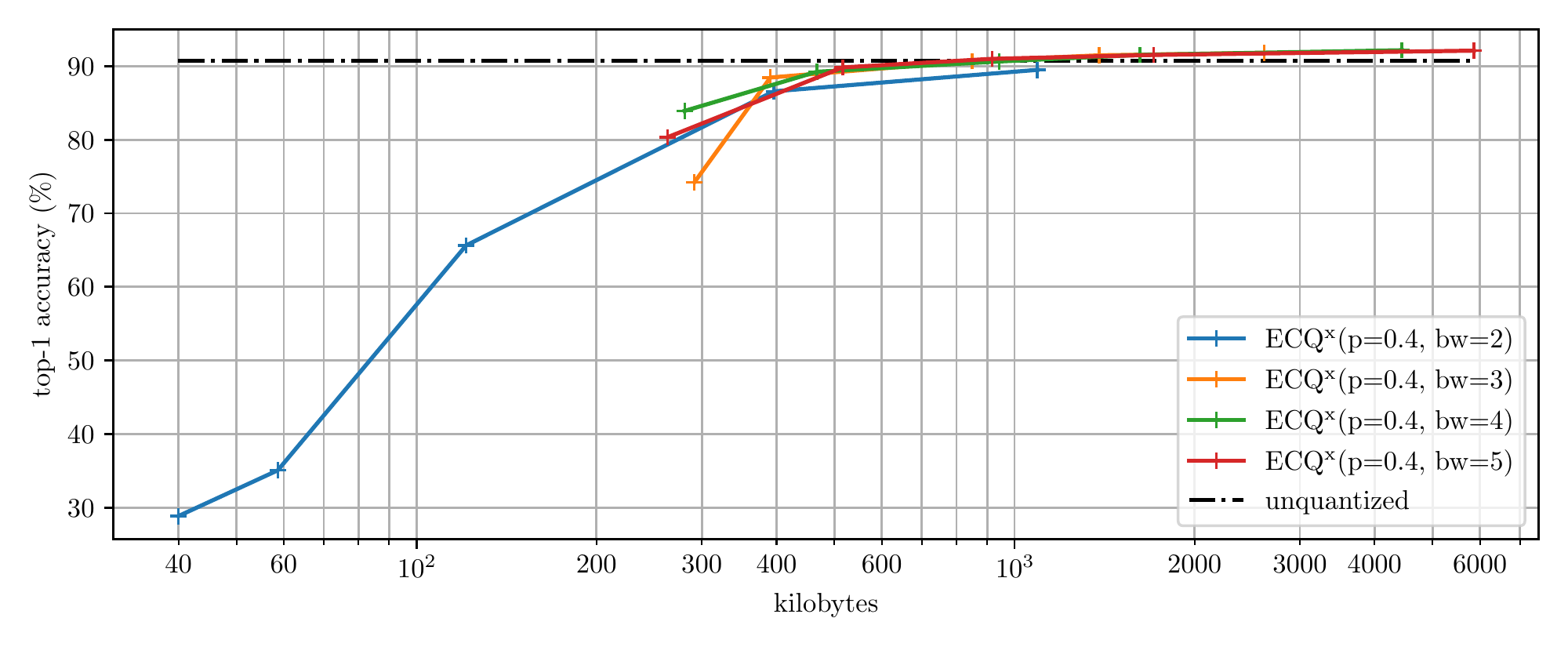}
\caption{Resulting VGG16 model performances vs. memory footprint, when applying ECQ$^{\text{x}}$ with 2 bit to 5 bit quantization.} \label{fig:bitwidth_vgg}
\end{figure}
Bit width reduction has multiple benefits over full-precision in terms of memory, latency, power consumption, and chip area efficiency.  For instance,  a reduction from standard 32 bit precision to 8 bit or 4 bit directly leads to a memory reduction of almost $4\times$ and $8\times$.
Arithmetic with lower bit width is exponentially faster if the hardware supports it.E.g., since the release of NVIDIA's Turing architecture,  4 bit integer is supported which increases the throughput of the RTX 6000 GPU to 522 TOPS (tera operations per second), when compared to 8 bit integer (261 TOPS) or 32 bit floating point (14.2 TFLOPS) \cite{nvidiaint4}.
Furthermore,  Horowitz showed that, for a 45\,nm technology, low-precision logic is significantly more efficient in terms of energy and area \cite{Horowitz}. For example, performing 8 bit integer addition and multiplication is 30$\times$ and 19$\times$ more energy efficient compared to 32 bit floating point addition and multiplication.  The respective chip area efficiency is increased by 116$\times$ and 27$\times$ as compared to 32 bit float.  It is also shown that memory reads and writes have the highest energy cost, especially when reading data from external DRAM. This further motivates bit width reduction because it can reduce the number of overall RAM accesses since more data fits into the same caches/registers when having a reduced precision.

In order to investigate different bit widths in the regime of ultra low precision,  we compare the compressibility and model performances of the MLP$\_$GSC and VGG16 networks when quantized to 2 bit, 3 bit, 4 bit and 5 bit integer values (see Figures~\ref{fig:bitwidth_mlp} and~\ref{fig:bitwidth_vgg}). Here, we directly encoded the integer tensors with the DeepCABAC codec of the ISO/IEC MPEG NNR standard \cite{NNROverview}. 
The least sparse working points of each trial, i.e., the rightmost data points of each line, show the expected behaviour, namely that compressibility is increased by continuously reducing the bit width from 5 bit to 2 bit. However, this effect decreases or even reverses when the bit width is in the range of 3 bit to 5 bit. In other words, reducing the number of centroids from $2^5=32$ to $2^3=8$ does not necessarily lead to a further significant reduction in the resulting bitstream size if sparsity is predominant. The 2 bit quantization still minimizes the size of the bit stream, even if, especially for the VGG model, more accuracy is sacrificed for this purpose. Note that compressibility is only one reason for reducing bit width besides, for example, speeding up model inference due to increased throughput.

\subsubsection{ECQ$^{\text{x}}$ Results Overview}
\begin{table}[!t]
{
\centering
\caption{Quantization results for ECQ$^{\text{x}}$ for 2 bit and 4 bit quantization: highest accuracy, highest compression gain without model degradation (if possible) and highest compression gain with negligible degradation. Underlined values mark the best results in terms of performance and compressibility with negligible drop in top-1 accuracy.}
\label{table:ecqx_vs_other}	
\resizebox{\linewidth}{!}{%
\begin{tabular}{lllcrcrr}
	\hline
	\bfseries Model & \bfseries Prec.$^a$ & \bfseries Method$^b$ & \bfseries Acc.~($\%$) &  \bfseries Acc. drop & \bfseries $\frac{|W=0|}{|W|}$~($\%$)$^c$ & \bfseries Size (kB) & \bfseries CR$^d$  \\
	\hline
	\hline \\
	\multicolumn{8}{c}{\bfseries CIFAR-10} \\
	\hline
	\hline
	\bfseries VGG16 & W4A16 & \bfseries ECQ$^{\text{x}}$ & 92.27 & \underline{+1.55} & 41.39 & 4,446.39 & 13.48 \\
	                & W4A16 & \bfseries ECQ$^{\text{x}}$ & 90.86 & +0.14 & 91.95 & 933.99 & 64.17 \\
	                & W4A16 &  \bfseries ECQ$^{\text{x}}$ & 90.62 & -0.10 & 94.67 & 584.16 & \underline{102.59} \\
	                & W4A16 & ECQ & 92.09 & +1.37 & 29.88 & 4,658.01 & 12.87 \\
	                & W4A16 & ECQ & 91.03 & +0.31 & 88.03 & 1,246.27 & 48.09 \\
	                & W4A16 & ECQ & 89.49 & -1.23 & 93.97 & 585.40 & 102.37 \\
	                \hline
	                & W2A16 & \bfseries ECQ$^{\text{x}}$ & 90.42 & \underline{-0.30} & 83.23 & 1.394,52 & \underline{42.98} \\
	                & W2A16 & ECQ & 90.19 & -0.53 & 81.58 & 1,486.76 & 40.31 \\
	\hline \\
	\multicolumn{8}{c}{\bfseries Google Speech Commands} \\
	\hline
	\hline
	\bfseries MLP$\_$GSC & W4A16 & \bfseries ECQ$^{\text{x}}$ & 88.95 & \underline{+0.71} & 65.14 & 128.03 & 20.05 \\
	                    & W4A16 & \bfseries ECQ$^{\text{x}}$ & 88.34 & +0.10 & 78.77 & 92.46 & 27.77 \\
	                     & W4A16 &\bfseries ECQ$^{\text{x}}$ & 87.89 & -0.34 & 80.45 & 87.52 & \underline{29.33} \\
	                     & W4A16 & ECQ & 88.71 & +0.47 & 59.95 & 139.96 & 18.34 \\
	                     & W4A16 & ECQ & 88.32 & +0.08 & 70.74 & 98.32 & 26.11\\
	                     & W4A16 & ECQ & 86.84 & -1.40 & 80.39 & 69.67 & 36.85 \\
	                     \hline
	                     & W2A16 &  \bfseries ECQ$^{\text{x}}$ & 87.46 & -0.78 & 83.97 & 68.77 & \underline{37.33} \\
	                     & W2A16 &  ECQ & 87.72 & \underline{-0.52} & 77.55 & 78.54 & 32.69 \\
	 \hline \\
	 \multicolumn{8}{c}{\bfseries Pascal VOC} \\
	 \hline
	 \hline
	 \bfseries ResNet18 & W4A16 & \bfseries ECQ$^{\text{x}}$ & 73.13 & \underline{-0.27} & 32.82 & 3,797.97 & 11.79 \\
	 & W4A16 & \bfseries ECQ$^{\text{x}}$ & 72.78 & -0.62 & 68.67 & 2,246.71 & \underline{19.93} \\
	 & W4A16 & \bfseries ECQ$^{\text{x}}$ & 72.48 & -0.92  & 74.65 & 1,946.22 & 23.01  \\
	 & W4A16 & ECQ & 72.95 & -0.45  & 24.63 & 3,882.62 & 11.53  \\
	 & W4A16 & ECQ & 72.56 & -0.84  & 61.12 & 2,480.59 & 18.05  \\
	 & W4A16 & ECQ & 71.74 & -1.66  & 74.88 & 1,841.82 & 24.32  \\
	\hline \\
\end{tabular}
}%
}
\footnotesize{$^{a}$ W$x$A$y$ indicates a quantization of weights and activations to $x$ and $y$ bit. } \\
\footnotesize{$^{b}$ ECQ refers to ECQ$^{\text{x}}$ w/o LRP constraint} \\ 
\footnotesize{$^{c}$ Sparsity, measured as the percentage of zero-valued parameters in the DNN.} \\
\footnotesize{$^{d}$ Compression ratio (full-precision size / compressed size) when applying the DeepCABAC codec of the ISO/IEC MPEG NNR standard \cite{NNROverview}.}
\end{table}
In addition to the performance graphs in the previous subsections, all quantization results are summarized in Table \ref{table:ecqx_vs_other}. Here, ECQ$^{\text{x}}$ and ECQ are compared specifically for a 2 and 4 bit quantization as these fit particularly well to power-of-two hardware registers. 
The ECQ$^{\text{x}}$ 4 bit quantization achieves a compression ratio for VGG16 of $103\times$ with a negligible drop in accuracy of $-0.1\%$. In comparison, ECQ achieves the same compression ratio only with a model degradation of $-1.23\%$ top-1 accuracy. For the 4 bit quantization of MLP\_GSC, ECQ$^{\text{x}}$ achieves its highest accuracy (``drop'', i.e., increase of $+0.71\%$ compared to the unquantized baseline model) with a compression ratio that is almost $10\%$ larger compared to the highest achievable accuracy of ECQ ($+0.47\%$). For sparsities beyond $70\%$, ECQ significantly reduces the model's predictive performance, e.g., at a sparsity of $80.39\%$ ECQ shows a loss of $-1.40\%$ whereas ECQ$^{\text{x}}$ only degrades by $-0.34\%$.
ResNet18 sacrifices performance at each quantization setting, but especially for ECQ$^{\text{x}}$ the accuracy loss is negligible. The 2 bit representations of ResNet18 sacrifice more than $-5\%$ top-1 accuracy compared to the unquantized model, which may be compensated with more than 20 epochs of quantization-aware training, but is also due to the higher complexity of the Pascal VOC task. 

And finally, the 2 bit results in Table \ref{table:ecqx_vs_other} show two major findings: 1) With only a minor model degradation all weight layers of the MLP\_GSC and VGG networks can also be quantized to only 4 discrete centroid values while still maintaining a high level of sparsity, 2) ECQ$^{\text{x}}$ renders higher compressible models in comparison to ECQ, as indicated by the higher compression ratios CR.

\section{Conclusion} \label{sec:conc}
In this chapter we presented a new entropy-constrained neural network quantization method (ECQ$^{\text{x}}$), utilizing weight relevance information from Layer-wise Relevance Propagation (LRP). Thus, our novel method combines concepts of explainable AI (XAI) and information theory. In particular, instead of only assigning weight values based on their distances to respective quantization clusters, the assignment function additionally considers weight relevances based on LRP. In detail, each weight's contribution to inference in interaction with the transformed data, as well as cluster information content is calculated and applied.
For this approach, we first utilized the observation that a weight's magnitude does not necessarily correlate with its importance or relevance for a model's inference capability.
Next, we verified this observation in a relevance vs. weight (magnitude) correlation analysis and subsequently introduce our ECQ$^{\text{x}}$ method.
As a result, smaller weight parameters that are usually omitted in a classical quantization process are preserved, if their relevance score indicates a stronger contribution to the overall neural network accuracy or performance.

The experimental results show that this novel ECQ$^{\text{x}}$ method generates low bit width (2-5 bit) and sparse neural networks while maintaining or even improving model performance. Therefore, in particular the 2 and 4 bit variants are highly suitable for neural network hardware adaptation tasks. Due to the reduced parameter precision and high number of zero-elements, the rendered networks are also highly compressible in terms of file size, e.g., up to $103\times$ compared to the full-precision unquantized DNN model, without degrading the model performance. Our ECQ$^{\text{x}}$ approach was evaluated on different types of models and datasets (including Google Speech Commands, CIFAR-10 and Pascal VOC). The comparative results vs. state-of-the-art entropy-constrained-only quantization (ECQ) show a performance increase in terms of higher sparsity, as well as a higher compression. Finally, also hyperparameter optimization and bit width variation results were presented, from which the optimal parameter selection for ECQ$^{\text{x}}$ was derived.

\section*{Acknowledgements}

This work was supported by the German Ministry for Education and Research as BIFOLD (ref.\ 01IS18025A and ref.\ 01IS18037A), the European Union’s Horizon 2020 programme (grant no.\ 965221 and 957059), and the Investitionsbank Berlin under contract No.\ 10174498 (Pro FIT programme).

\bibliographystyle{splncs04}
\bibliography{main_20220216210608}

\end{document}